\documentclass{article}
\usepackage{graphicx}
\usepackage{amsmath}
\usepackage{cite}
\usepackage{url}

\begin{document}

\title{Shape Complexity in Cluster Analysis}

\author{
Eduardo~J.~Aguilar\\
Instituto de Ci\^encia e Tecnologia\\
Universidade Federal de Alfenas\\
Rod.\ Jos\'e Aur\'elio Vilela, 11999\\
37715-400 Po\c cos de Caldas - MG, Brazil\\
\\
Valmir~C.~Barbosa\thanks{Corresponding author (valmir@cos.ufrj.br).}\\
Programa de Engenharia de Sistemas e Computa\c c\~ao, COPPE\\
Universidade Federal do Rio de Janeiro\\
Centro de Tecnologia, Sala H-319\\
21941-914 Rio de Janeiro - RJ, Brazil}

\date{}

\maketitle

\begin{abstract}
In cluster analysis, a common first step is to scale the data aiming to better
partition them into clusters. Even though many different techniques have
throughout many years been introduced to this end, it is probably fair to say
that the workhorse in this preprocessing phase has been to divide the data by
the standard deviation along each dimension. Like division by the standard
deviation, the great majority of scaling techniques can be said to have roots in
some sort of statistical take on the data. Here we explore the use of
multidimensional shapes of data, aiming to obtain scaling factors for use prior
to clustering by some method, like k-means, that makes explicit use of distances
between samples. We borrow from the field of cosmology and related areas the
recently introduced notion of shape complexity, which in the variant we use is
a relatively simple, data-dependent nonlinear function that we show can be used
to help with the determination of appropriate scaling factors. Focusing on what
might be called ``midrange'' distances, we formulate a constrained nonlinear
programming problem and use it to produce candidate scaling-factor sets that
can be sifted on the basis of further considerations of the data, say via expert
knowledge. We give results on some iconic data sets, highlighting the strengths
and potential weaknesses of the new approach. These results are generally
positive across all the data sets used.

\end{abstract}

\newpage
\section{Introduction}
\label{intr}

The common wisdom regarding the processing of data prior to cluster analysis,
particularly when a distance-based clustering method like k-means or some forms
of hierarchical clustering are used, is that data should be scaled to improve
results. Even though researchers have been prolific in creating domain-specific
forms of scaling (cf., e.g., \cite{bhwsw06}), already the earliest studies
systematically approaching the subject viewed division by the standard deviation
or by the range in each dimension as the natural candidates they still are to
this day \cite{e79,mc88}. This is not to say that alternative divisors were not
considered: they were \cite{s04}, but the situation seems to have remained
largely unchanged until very recently, with the introduction of the so-called
pooled standard deviation \cite{rz20}, which continues to support division by
the standard deviation unless this would make the dimension in question lose
information crucial to partitioning the data into clusters. Should this be the
case, a weighted averaged of the standard deviations localized around
statistically significant modes in that dimension is used instead. This average
is the pooled standard deviation of the data in that dimension, henceforth
denoted by $\sigma_k^\mathrm{pool}$ for dimension $k$. Notably, the essential
motivation for the creation of $\sigma_k^\mathrm{pool}$ seems well aligned with
concerns that have been voiced since the late 1960's (cf.\ \cite{mc88} for
comments on this).

In a similar vein, it has for several decades been clear that some form of
optimization problem must exist whose solution yields scale factors that make
some sort of sense for the various dimensions. And indeed this has been pursued,
though to the best of our knowledge not for the last three decades, at least.
Noteworthy representatives of these attempts include optimizing for a linear
transformation of the data \cite{k72}; maximizing the square of a correlation
between two sets of distances between samples \cite{dcc84}; a least-squares
method for determining scale factors that make such distances approach those
in the dendrogram resulting from hierarchical clustering \cite{sdc85}; and
determining scaling factors by considering the modal structure of the data in
each dimension in a way that, to a certain degree, prefigures the above
definition of the pooled standard deviation \cite{h86}. Each of these approaches
seems to have either disappointed its own creator \cite{k72}, or remained
tailored to the generally uninteresting cases of nonoverlapping clusters
\cite{dcc84}, or been tested only superficially \cite{sdc85}, or simply remained
untested \cite{h86}.

Here we introduce the use of a shape-complexity function to guide the
determination of scale factors. By this denomination we are not referring to one
of the many forms of complexity used to characterize the computer
representations of three-dimensional shapes \cite{r05}. Instead, we refer to a
generalization to multiple dimensions of the homonymous three-dimensional
concept introduced recently in cosmology and related disciplines \cite{m18,b20}.
If we imagine (up to three dimensions) that the disposition of data samples in
space gives the data set an inherent shape, then clearly being able to shrink or
stretch each dimension independently of all others is an important source of
shape variation, one that we explore for the purpose of cluster analysis. One
particular facet of shape complexity that we find especially relevant to this
end is that it allows what should be intercluster distances to be considered
side-by-side with distances that should be intracluster. While normally cluster
analysts know which distance is which type only in a very limited manner, shape
complexity provides a handle that can help in posing a nonlinear programming
problem for the automatic determination of scale factors (or rather, candidate
scale factors to undergo further scrutiny based on what analysts do know of the
domain in question).

We proceed in the following manner. In Section~\ref{scompl} we introduce the
form of shape complexity we use, giving its definition and properties of
interest. We also discuss why it relates closely to the role of scale factors
in cluster analysis and how determining such factors from it can be formulated.
We then proceed with a description of our experimental setup, in
Section~\ref{exp}. This includes the data sets on which we experiment, the
tools and algorithms we use, and how we evaluate a data set's partition into
clusters. Importantly, we perform clustering solely via the k-means method,
owing mainly to its great potential to perform well when clusters overlap
\cite{fs18}, and also to its long history, during which many implementations and
variants have appeared \cite{j10}. Results and discussion are given in
Sections~\ref{res} and~\ref{disc}, respectively. We close in
Section~\ref{concl}.

\section{Shape complexity}
\label{scompl}

We consider an $n_\mathrm{orig}\times d$ data matrix $X$ with
$n_\mathrm{orig},d>1$, where $n_\mathrm{orig}$ is the number of $d$-dimensional
real samples. For $1\le k\le d$, we use $\sigma_k^2$ to denote the samples'
variance on dimension $k$, and $\alpha_k>0$ to denote the scale factor to be
used on the samples along this dimension in order to facilitate clustering.
Factor $\alpha_k$ is assumed to be applied to the various $X_{ik}$'s along with
their division by $\sigma_k$. That is, each original $X_{ik}$ is to undergo
scaling by the factor $\alpha_k/\sigma_k$. This makes it easier to assess the
effect of factor $\alpha_k$ relative to the more common $1/\sigma_k$ and also
enables some key developments later in this section.

For reasons to be discussed shortly, here we propose that the appropriate
$\alpha_k$'s be determined with the guidance of the so-called shape complexity
of the $n_\mathrm{orig}$ points in $d$-dimensional real space that define the
samples. This notion is borrowed from the physics of multiple bodies interacting
gravitationally with one another. For $d=3$ and the points having masses
associated with them, shape complexity has been shown to help account for
structure as it arises in the form of clusters during the system's evolution
\cite{m18,b20}.

The version of shape complexity we use, denoted by $\mathrm{SC}$, is given by
\begin{equation}
\mathrm{SC}=
\biggl(\sum_{i<j}r_{ij}^2\biggr)^{1/2}\sum_{i<j}r_{ij}^{-1},
\label{sc}
\end{equation}
where each $i$ and $j$ are distinct samples and $r_{ij}$ is the Euclidean
distance between them. That is, the number of samples $\mathrm{SC}$ takes into
account is $n$ such that $1<n\leq n_\mathrm{orig}$ (duplicates may thus exist
only if $n_\mathrm{orig}>2$)
and
\begin{equation}
r_{ij}^2=\sum_k\alpha_k^2\rho_{ijk}^2,
\label{rij}
\end{equation}
with
\begin{equation}
\rho_{ijk}=\frac{X_{ik}-X_{jk}}{\sigma_k}.
\end{equation}

$\mathrm{SC}$ is therefore a function of the $\alpha_k$'s, but we refrain from
denoting this explicitly for the sake of notational clarity. Importantly, the
use of $i<j$ in the summations of Eq.~(\ref{sc}) indicates that they occur on
the set of all $\binom{n}{2}$ unordered pairs of distinct samples. Likewise, the
summation on $k$ in Eq.~(\ref{rij}) indicates that it occurs over all $d$
dimensions.

\paragraph{Radial invariance.}
One of the key properties for which $\mathrm{SC}$ is appreciated in its fields
of origin is scale invariance, which in our terms is to be understood as
follows. If $\alpha_k$ has the same value for every $k$, then clearly
$\mathrm{SC}$ remains unchanged however this common value is varied. But if
$\mathrm{SC}$ is to be used to improve the results of clustering algorithms on
the data, setting every $\alpha_k$ to the same value is in general not an
option. Scale invariance, nevertheless, is a special case of the much more
useful radial invariance we discuss next. The radial invariance of $\mathrm{SC}$
can be seen in more than one way, but here we choose the perspective of certain
directional derivatives of $\mathrm{SC}$. This requires us to already consider
the gradient of $\mathrm{SC}$, which will be instrumental later on.

For $f$ a differentiable function of the $\alpha_k$'s, we let $f'_k$ denote
$\partial f/\partial\alpha_k$, the $k$th component of the gradient of $f$, and
moreover write $g=(\sum_{i<j}r_{ij}^2)^{1/2}$ and $h=\sum_{i<j}r_{ij}^{-1}$ so
that $\mathrm{SC}=gh$. We get
\begin{equation}
\mathrm{SC}'_k=g'_kh+gh'_k,
\end{equation}
where
\begin{equation}
g'_k=\alpha_kg^{-1}\sum_{i<j}\rho_{ijk}^2
\end{equation}
and
\begin{equation}
h'_k=-\alpha_k\sum_{i<j}r_{ij}^{-3}\rho_{ijk}^2.
\end{equation}

Radial invariance comes from realizing that the directional derivative of
$\mathrm{SC}$ is zero along any straight line extending out from the origin into
the positive $d$-dimensional real orthant, that is,
\begin{equation}
\sum_k\alpha_k\mathrm{SC}'_k=0
\label{rdi}
\end{equation}
for any valuation of the $\alpha_k$'s. To put it differently, $\mathrm{SC}$ has
the same value at any two assignments of values to $\alpha_1,\ldots,\alpha_d$,
say $v_1^{(1)},\ldots,v_d^{(1)}$ and $v_1^{(2)},\ldots,v_d^{(2)}$, such that
$v_k^{(2)}=tv_k^{(1)}$ for every $k$ and some $t>0$.

To see how Eq.~(\ref{rdi}) comes about, simply write
\begin{align}
g\sum_k\alpha_k\mathrm{SC}'_k
&=
h\sum_k\alpha_k^2\sum_{i<j}\rho_{ijk}^2-
g^2\sum_k\alpha_k^2\sum_{i<j}r_{ij}^{-3}\rho_{ijk}^2\\
&=
h\sum_{i<j}\sum_k\alpha_k^2\rho_{ijk}^2-
g^2\sum_{i<j}r_{ij}^{-3}\sum_k\alpha_k^2\rho_{ijk}^2\\
&=
h\sum_{i<j}r_{ij}^2-g^2\sum_{i<j}r_{ij}^{-3}r_{ij}^2\\
&=
hg^2-g^2h.
\end{align}
The role of Eq.~(\ref{rij}) in this development highlights a condition
equivalent to radial invariance: that the value of any $r_{ij}$ becomes scaled
by $t$ when moving from $v_1^{(1)},\ldots,v_d^{(1)}$ to
$v_1^{(2)},\ldots,v_d^{(2)}$.

\paragraph{Shape complexity and clustering.}
Increasing any $r_{ij}$ always increases $g$ while decreasing $h$. Notably, the
most significant increases in $g$ come from increasing the largest $r_{ij}$'s
(since $\partial g/\partial r_{ij}=g^{-1}r_{ij}$), while the most significant
decreases in $h$ come from increasing the smallest $r_{ij}$'s (since
$\partial h/\partial r_{ij}=-r_{ij}^{-2}$). Because increases in the $r_{ij}$'s
are mediated by increases in the $\alpha_k$'s, the effect of increasing any
specific $\alpha_k$ on the $r_{ij}$'s of specific relative magnitudes is best
understood by considering how the ratios $g'_k/g$ and $-h'_k/h$ relate to each
other. Two cases must be considered, as follows.

\begin{enumerate}
\item[C1.] If $g'_k/g>-h'_k/h$ (i.e., increasing $\alpha_k$ causes more of a
relative increase in $g$ than a relative decrease in $h$), then larger
$r_{ij}$'s are being increased more than smaller $r_{ij}$'s.
\item[C2.] If $g'_k/g<-h'_k/h$ (i.e., increasing $\alpha_k$ causes more of a
relative decrease in $h$ than a relative increase in $g$), then smaller
$r_{ij}$'s are being increased more than larger $r_{ij}$'s.
\end{enumerate}

In the context of data clustering, assume for a moment that larger $r_{ij}$'s
are generally intercluster distances while smaller $r_{ij}$'s are generally
intracluster distances. Cases C1 and C2 above are then in strong opposition to
each other, as clearly case C1 could be good for clustering and case C2 bad for
clustering. We might then expect to be well-off if we targeted case C1 for every
$k$, but surely an assignment of values to $\alpha_1,\ldots,\alpha_d$ might
satisfy case C1 for a specific $k$ while satisfying case C2 for another. In this
case it would seem better to pursue the intermediate goal of getting as close as
possible to achieving $g'_k/g=-h'_k/h$ for every $k$.

Real-world data, however, rarely comply with the dichotomy we momentarily
assumed above. Instead, quite often larger distances are intracluster, and
likewise smaller distances are intercluster. In any case, the centerpiece of the
strategy we adopt henceforth is the same that would be appropriate had the
dichotomy always held true, that is, seeking the equilibrium represented by
$g'_k/g=-h'_k/h$ for every $k$. On top of this, we essentially look for several
scaling-factor schemes approaching such conditions as closely as possible and
select the one (or more than one) that upon closer inspection of the data leads
to a reasonable partition into clusters.

\paragraph{The optimization problem.}
A consequence of our discussion of the radial-invariance property of
$\mathrm{SC}$ is that all assignments of values to $\alpha_1,\ldots,\alpha_d$
on any straight line emanating from the origin into the positive
$d$-dimensional real orthant are equivalent at providing scaling factors for
distance-based clustering. That is, choosing any such assignment will lead any
distance-based clustering algorithm to yield the same result. This follows from
the fact that the $r_{ij}$'s for a given assignment on that line are scaled
versions, by the same factor on all dimensions, of $r_{ij}$'s for any of the
other assignments.

In what follows, all but one of such equivalent assignments are ignored. The one
that is taken into account is that for which
\begin{equation}
\sum_k\alpha_k^2=d.
\label{ec}
\end{equation}
That is, valid assignments of values to the $\alpha_k$'s must be on the
$d$-dimensional sphere of radius $\sqrt d$ centered at the origin. This choice
of radius allows for $\alpha_k=1$ for every $k$ to be a valid assignment. This,
we recall from earlier in this section, is the assignment that scales the data
along dimension $k$ by the factor $1/\sigma_k$.

Seeking to approximate $g'_k/g=-h'_k/h$ for every $k$ given this equality
constraint boils down to the problem of finding a local minimum or maximum of
$\mathrm{SC}$ given the constraint. Because $\mathrm{SC}$ is inextricably based
on the data to be clustered, it seems to have no characteristic that can be
directly exploited to this end. We follow an indirect route and begin by
considering the first-order necessary condition for local optimality in this
case \cite{l73}, which requires not only the equality constraint in
Eq.~(\ref{ec}) to be satisfied but also the gradient of the corresponding
Lagrangian with respect to the $\alpha_k$'s to equal zero. The Lagrangian in
this case is
\begin{equation}
L=\mathrm{SC}+\lambda(\sum_k\alpha_k^2-d),
\end{equation}
where $\lambda$ is the Lagrange multiplier corresponding to the single equality
constraint. Its gradient's $k$th component is
$L'_k=g'_kh+gh'_k+2\lambda\alpha_k$. Writing this in more detail yields
\begin{equation}
L'_k=
\alpha_k\biggl(
g^{-1}h\sum_{i<j}\rho_{ijk}^2-
g\sum_{i<j}r_{ij}^{-3}\rho_{ijk}^2+
2\lambda
\biggr),
\end{equation}
from which it follows that, in order to achieve $L'_k=0$ for every $k$, we must
have
\begin{equation}
\sum_{i<j}r_{ij}^{-3}\frac{\rho_{ijk}^2}{\sum_{i<j}\rho_{ijk}^2}=
\frac{g^{-1}h+2\lambda(\sum_{i<j}\rho_{ijk}^2)^{-1}}{g}
\label{dtpr}
\end{equation}
for each of them.

Even though it would seem that the right-hand side of Eq.~(\ref{dtpr}) may have
a different value for each $k$, letting $N=n_\mathrm{orig}(n_\mathrm{orig}-1)$
we note that $\sigma_k^2$ can be written as
\begin{equation}
\sigma_k^2=N^{-1}\sum_{i<j}(X_{ik}-X_{jk})^2,
\end{equation}
which leads to
\begin{equation}
\sum_{i<j}\rho_{ijk}^2=N.
\label{sumN}
\end{equation}
The right-hand side of Eq.~(\ref{dtpr}) is therefore the same for every $k$, so
its left-hand side, which is in fact the inner product of two vectors in
$\binom{n}{2}$-dimensional real space, must also not depend on $k$. The $ij$th
component of one of the two vectors involved in this inner product is
$N^{-1}\rho_{ijk}^2$, so all $d$ such vectors are coplanar, since by
Eq.~(\ref{sumN}) they all lie on the $\binom{n}{2}$-dimensional plane
$\sum_{i<j}x_{ij}=1$. In order for the inner product to have the same value
regardless of $k$, the vector of $ij$th component $r_{ij}^{-3}$ must therefore
be orthogonal to this plane. That is, we must have
\begin{equation}
\sum_{i<j}r_{ij}^{-3}N^{-1}(\rho_{ijk}^2-\rho_{ij\ell}^2)=0,
\label{grad0}
\end{equation}
where $k,\ell$ are any two of the $d$ dimensions. In the formulation that
follows we use $k=1$, $\ell=2$.

Determining the $\alpha_k$'s directly from Eq.~(\ref{grad0}) and the equality
constraint is not possible, so we resort to the following nonlinear programming
problem instead.
\begin{align}
\text{minimize }
&\textstyle
\bigl(\sum_{i<j}r_{ij}^{-3}N^{-1}(\rho_{ij1}^2-\rho_{ij2}^2)\bigr)^2
\label{obj}\\
\text{subject to }
&\textstyle\sum_k\alpha_k^2=d,
\label{c1}\\
&\alpha_k>0.&\forall k\in\{1,\ldots,d\}
\label{c2}
\end{align}
Solving this problem will return $\alpha_k$'s that approximate Eq.~(\ref{grad0})
as well as possible. Even if a good approximation is returned, it must be kept
in mind that only first-order necessary conditions are being taken into account.
The second-order necessary and sufficient conditions, which involve the second
derivatives of $\mathrm{SC}$, are not. Further methodological steps must then be
taken, as detailed in Section~\ref{exp} along with the necessary tool set.
Henceforth, we refer to the optimization problem given in
Eqs.~(\ref{obj})--(\ref{c2}) simply as Problem~P.

\paragraph{Further remarks.}
Another consequence of Eq.~(\ref{sumN}) is that
\begin{equation}
\sum_{i<j}r_{ij}^2=N\sum_k\alpha_k^2,
\label{sumr2}
\end{equation}
which allows $g$ to be rewritten as
\begin{equation}
g=\biggl(N\sum_k\alpha_k^2\biggr)^{1/2}.
\end{equation}

\section{Experimental setup}
\label{exp}

Solving Problem~P to discover the $\alpha_k$'s is the centerpiece of our
approach. Several candidate sets of these scaling factors can be obtained by
solving the problem repeatedly in a sequence of random trials, each one first
selecting an initial point for the minimization and then attempting to converge
to a set of $\alpha_k$'s for which the problem's objective function is locally
minimum. The resulting scaling-factor sets can then be pitched against one
another, engaging the user's knowledge of the data set for the selection of a
small set of candidates (perhaps even a single one) to carry on with.

Because clustering is an approach to data analysis that depends strongly on
a domain expert's knowledge of and familiarity with the data set, uncertainties
during the process of selecting appropriate scaling-factor sets from those
turned up by solving Problem~P are inevitable. To illustrate some strategies to
deal with this, in Sections~\ref{res} and~\ref{disc} we discuss our experience
with analyzing five well-known benchmarks in light of $\mathrm{SC}$. Dealing
with these data sets has of course been greatly facilitated by the availability
of the reference partition into clusters for each one. This will not be
available in a real-world scenario, except perhaps in some fragmentary form, but
in our discussion of the benchmarks we attempt to provide viewpoints that may be
useful even then.

We continue this section with the presentation of the benchmarks we use, and of
the tools, algorithms, and evaluation method we enlist.

\paragraph{Data sets.}
The five data sets we use are listed in Table~\ref{tab1}, along with crucial
information on them. We divide them into two groups, based on our experience in
handling them, particularly on the difficulty in obtaining good partitions. The
first group contains those for which it has proven possible to obtain partitions
that approximate the corresponding reference partition well. The second group
contains those for which approximating the reference partition, even if only
reasonably, has proven harder.

\begin{table}[t]
\centering
\caption{Data sets used and their properties.}
\label{tab1}
\makebox[\textwidth][c]{\begin{tabular}{lllll}
\hline
Data set, & Number of & Number of & Number of & Number of \\
number of & samples & unique & actual/original & dimensions with \\
clusters & ($n_\mathrm{orig}$) & samples & dimensions &missing values \\
& & ($n$) & ($d$/$d_\mathrm{orig}$) & ($d_\mathrm{miss}$) \\
\hline
Iris, 3 & 150 & 149 & 4/4 & 0 \\
BCW, 2 & 699 & 465 & 9/9 & 1 \\
BC-DR3, 4 & 62 & 62 & 3/496 & 392 \\
BNA-DR3, 2 & 1372 & 1348 & 3/4 & 0 \\
BCW-Diag-10, 2 & 569 & 569 & 10/30 & 0 \\
\hline
\end{tabular}
}
\end{table}

The first group has two members, the Iris data set (downloaded from
\cite{urlIris} and then corrected to exactly match the data in the original
publication \cite{f36}) and BCW, the original version of the Wisconsin breast
cancer data sets \cite{urlBCW}.

The second group comprises three data sets, viz.: BC-DR3, which comes from the
version of Perou et al.'s breast cancer data set \cite{pserjrprjafpwzlbbb00}
compiled and made available by the proponents of scaling by the
$\sigma_k^\mathrm{pool}$'s mentioned in Section~\ref{intr}; BNA-DR3, from a data
set containing wavelet-transform versions and the entropy of banknote images for
authentication \cite{urlBNA}; and BCW-Diag-10, from the so-called diagnostic
version of the Wisconsin breast cancer data sets \cite{urlBCWDiag}.

The three data sets in the second group have fewer dimensions than originally
available, which is indicated in Table~\ref{tab1} by the $d<d_\mathrm{orig}$
values on the fourth column. We reduced these data sets' numbers of dimensions
as an attempt to make clustering succeed better than it would otherwise. In two
cases this is indicated by the ``DR3'' in the data sets' names, which refers to
dimensionality reduction by adopting the first three principal components output
by principal component analysis (PCA) \cite{jc16} on the data after centering
(but not scaling) the samples. This was done in the R language, using function
\verb|prcomp| with options \verb|center = T| and \verb|scale = F|. The resulting
BC-DR3 and BNA-DR3 retain 35.85\% and 97.02\% of the original variance,
respectively.

The third case is that of BCW-Diag-10, which contains only the first 10 of the
original 30 dimensions. Each sample in this data set is an image and each
dimension is a statistic computed on that image. The 10 dimensions we use are
mean values.

As per the second and third columns in the table, three of the data sets (Iris,
BCW, and BNA-DR3) contain more samples ($n_\mathrm{orig}$) than unique samples
($n$). The difference corresponds to duplicates, which were discarded so that
$\mathrm{SC}$, and consequently Problem~P, could be defined properly. The
table's fifth column is also worthy of attention, since it gives for each data
set the number of dimensions ($d_\mathrm{miss}$) for which missing values are to
be found. Such values were synthesized in the Wolfram Mathematica 13.0 system,
using function \verb|SynthesizeMissingValues| with default settings. This caused
no further duplicates to appear and, in the case of BC-DR3, was of course done
before PCA.

\paragraph{Computational tools and algorithms.}
For each of the data sets in Table~\ref{tab1}, first we ran $1\,000$ trials,
each one aiming to obtain a scaling-factor candidate set by solving Problem~P.
We did the required optimization in the Wolfram Mathematica 13.0 system, using
function \verb|FindMinimum|, mostly with default settings, to find a local
minimum of the problem's objective function for each trial. Our only choices of
a non-default setting for \verb|FindMinimum| were the following: for each trial
we specified an initial point in $[0.5,1.5]^d$, selected uniformly at random by
an application of function \verb|RandomReal| to each dimension; we allowed for a
larger number of iterations with the \verb|MaxIterations -> 5000| option; and we
precluded any symbolic manipulation with \verb|Gradient -> "FiniteDifference"|
(because Problem~P is strongly data-dependent, allowing Mathematica to perform
the symbolic manipulations that come so naturally to it can quickly lead to
memory overflow).

On occasion we have noticed that coding the constraint in Eq.~(\ref{c2}) as is
can lead to division-by-zero errors. We thus avoided this by expressing the
constraint as $\alpha_k\geq 10^{-5}$ instead of $\alpha_k>0$. Still regarding
errors during minimization, \verb|FindMinimum| can also fail by not attaining
convergence within the specified maximum number of iterations. This error can
take more than one form, but in general we have observed it in no more than
0.2\% of the trials for each data set. When a failure of this type does occur a
solution is still output, but in our experiments such outputs were discarded
when compiling results.

One crucial step in this study is of course partitioning the data into clusters
after they have been appropriately scaled. We performed this step in the R
language, using in all cases the k-means method as implemented in function
\verb|kmeans|. Because k-means has certain randomized components, we first set a
fixed seed, via \verb|set.seed(1234)|, to facilitate consistency checks.
Function \verb|kmeans| receives as input the scaled version of the
$n_\mathrm{orig}\times d$ data matrix $X$ and also the desired number of
clusters (the same as in the data set's reference partition). For the results we
report in Section~\ref{res}, scaling happened according to one of four
possibilities: either each $X_{ik}$ remained unchanged (no scaling), or it
became one of $(1/\sigma_k)X_{ik}$, $(1/\sigma_k^\mathrm{pool})X_{ik}$, or
$(\alpha_k/\sigma_k)X_{ik}$. The latter is scaling as indicated by some random
trial with Problem~P.

\paragraph{Partition evaluation.}
To evaluate the partition resulting from clustering we use the variant of the
well-known Adjusted Rand Index (ARI) \cite{ha85} that seems most appropriate in
our present context, which is that every possible resulting partition of a data
set by the clustering algorithm in use must have a fixed number of clusters
(``$\mathrm{fnc}$,'' used in notations henceforth) \cite{ga17}. That this is
clearly the case follows from our use of k-means described above. The ARI
variant is
\begin{equation}
\mathrm{ARI_{fnc}}=
\frac{\mathrm{RI}-\mathrm{E_{fnc}[RI]}}{1-\mathrm{E_{fnc}[RI]}},
\end{equation}
where $\mathrm{RI}$ is the original Rand Index and $\mathrm{E_{fnc}[RI]}$ is its
expected value given the fixed number of clusters condition.

Letting $C$ denote the number of clusters that any obtained partition will have,
the formulas for $\mathrm{RI}$ and $\mathrm{E_{fnc}[RI]}$ are
\begin{equation}
\mathrm{RI}=\binom{n_\mathrm{orig}}{2}^{-1}(\mathrm{TS}+\mathrm{TD})
\label{ri}
\end{equation}
and
\begin{equation}
\mathrm{E_{fnc}[RI]}= UV+(1-U)(1-V),
\end{equation}
with
\begin{equation}
U=
\genfrac\{\}{0pt}{0}{n_\mathrm{orig}}{C}^{-1}
\genfrac\{\}{0pt}{0}{n_\mathrm{orig}-1}{C}
\label{u}
\end{equation}
and
\begin{equation}
V=
\binom{n_\mathrm{orig}}{2}^{-1}
(\mathrm{TS}+\mathrm{FD}).
\label{v}
\end{equation}

In Eqs.~(\ref{ri}) and~(\ref{v}), $\mathrm{TS}$ (for ``true similar'') counts
the number of sample pairs that are in the same cluster according to the
obtained partition and in the same cluster according to the reference partition;
$\mathrm{TD}$ (``true dissimilar'') counts pairs that are split between
different clusters according to both the obtained partition and the reference
partition; and $\mathrm{FD}$ (``false dissimilar'') counts those that are split
between different clusters according to the obtained partition but are in the
same cluster according to the reference partition. Curly brackets are used in
Eq.~(\ref{u}) to denote Stirling numbers of the second kind.

$\mathrm{ARI_{fnc}}$ equals at most $1$, which happens for
$\mathrm{TS}+\mathrm{TD}=\binom{n_\mathrm{orig}}{2}$ (i.e., when the obtained
partition and the reference partition are identical).

\section{Results}
\label{res}

All our results are in reference to the data sets listed in Table~\ref{tab1}
and are summarized in Table~\ref{tab2}. In this table, the value of
$\mathrm{ARI_{fnc}}$ resulting from the use of k-means is given for each of
several scaled versions of the $n_\mathrm{orig}\times d$ data matrix $X$ that
corresponds to each data set. There are four schemes in each case: the
no-scaling scheme, in which each $X_{ik}$ is used directly as it appears in the
data matrix; the scheme that makes use of the standard deviation in each
dimension, in which $X_{ik}$ is scaled by $1/\sigma_k$; the scheme that uses
pooled standard deviations instead, in which $X_{ik}$ is scaled by
$1/\sigma_k^\mathrm{pool}$; and the scheme that uses scaling factors obtained by
solving Problem~P, in which $X_{ik}$ is scaled by $\alpha_k/\sigma_k$ for the
resulting $\alpha_k$.

\begin{table}[t]
\centering
\caption{Performance of k-means, according to $\mathrm{ARI_{fnc}}$, on various
scaled versions of the data sets in Table~\ref{tab1}.}
\label{tab2}
\makebox[\textwidth][c]{\begin{tabular}{lllll}
\hline
Data set & \multicolumn{4}{c}{$\mathrm{ARI_{fnc}}$} \\
\cline{2-5}
& No scaling & Scaling by & Scaling by & Scaling by \\
& & $1/\sigma_k$ & $1/\sigma_k^\mathrm{pool}$ & $\alpha_k/\sigma_k$ \\
\hline
Iris & 0.728 & 0.621 & 0.886 & 0.571--0.904 \\
BCW& 0.840 & 0.825 & 0.825 & 0.189--0.877 \\
BC-DR3 & 0.492 & 0.518 & 0.518 & ($-0.021$)--0.535 \\
BNA-DR3 & 0.050 & 0.023 & 0.023 & ($-0.000$)--0.659 \\
BCW-Diag-10 & 0.456 & 0.673 & 0.673 & 0.633--0.655 \\
\hline
\end{tabular}
}
\end{table}

$\mathrm{ARI_{fnc}}$ values for the latter type of scaling are presented in
Table~\ref{tab2} as intervals, indicating in each case the lowest and the
highest value observed in the $1\,000$ random trials with Problem~P (slightly
fewer trials if optimization errors happened). For BC-DR3 and BNA-DR3, the
intervals begin at slightly negative values of $\mathrm{ARI_{fnc}}$ (indicating
that $\mathrm{RI}<\mathrm{E_{fnc}[RI]}$).

The intervals on the rightmost column of Table~\ref{tab2} are supplemented by
the panels in Figure~\ref{cand}, where each row of panels (rows A through E)
corresponds to one of the data sets. Such panels allow viewing the various
$\mathrm{ARI_{fnc}}$ values inside those intervals from different perspectives.
The left panel on each row is a plot of $\mathrm{SC}$ against $\alpha_1$ for all
trials on the corresponding data set. The choice of $\alpha_1$ is completely
arbitrary and meant only to offer a glimpse into how $\mathrm{SC}$ depends on
the $\alpha_k$'s turned up by solving Problem~P. Points are color-coded to
indicate how their $\mathrm{ARI_{fnc}}$ values relate to one another. The right
panel on each row allows viewing such values as a histogram.

\begin{figure}[p]
\centering
\makebox[\textwidth][c]{\includegraphics[scale=0.35]{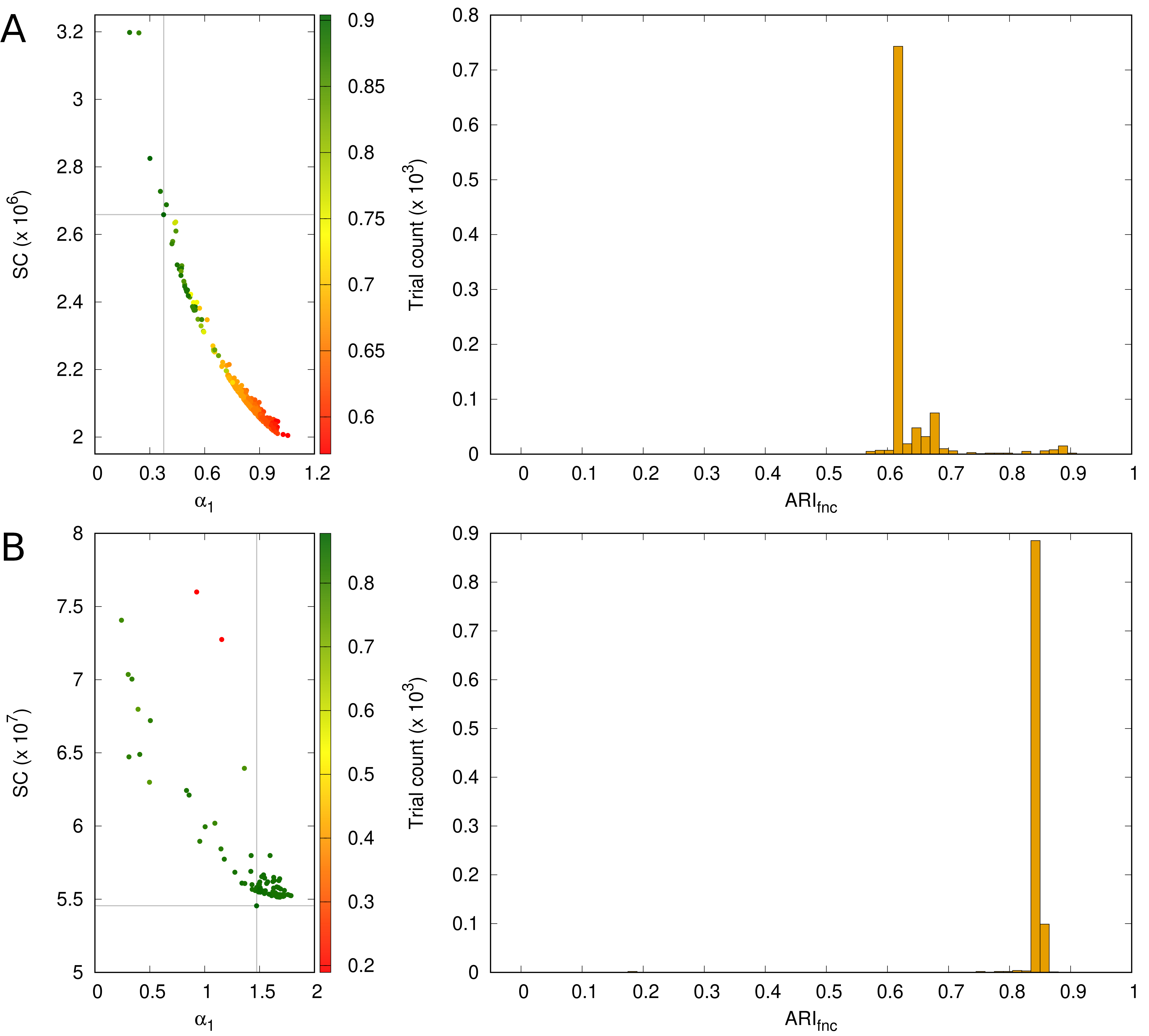}}
\caption{Results of the random trials with Problem~P on Iris (A), BCW (B),
BC-DR3 (C), BNA-DR3 (D), and BCW-Diag-10 (E), expanding on the summary given on
the rightmost column of Table~\ref{tab2}. Each point on each left panel
corresponds to a trial and is color-coded according to the accompanying palette
to reflect the value of $\mathrm{ARI_{fnc}}$ it leads to by way of clustering
with k-means. The point leading to the highest $\mathrm{ARI_{fnc}}$ value is
marked by the crosshair in the panel. Each right panel provides a view of how
$\mathrm{ARI_{fnc}}$ is distributed over all pertaining trials.}
\label{cand}
\end{figure}

\addtocounter{figure}{-1}
\begin{figure}[p]
\centering
\makebox[\textwidth][c]{\includegraphics[scale=0.35]{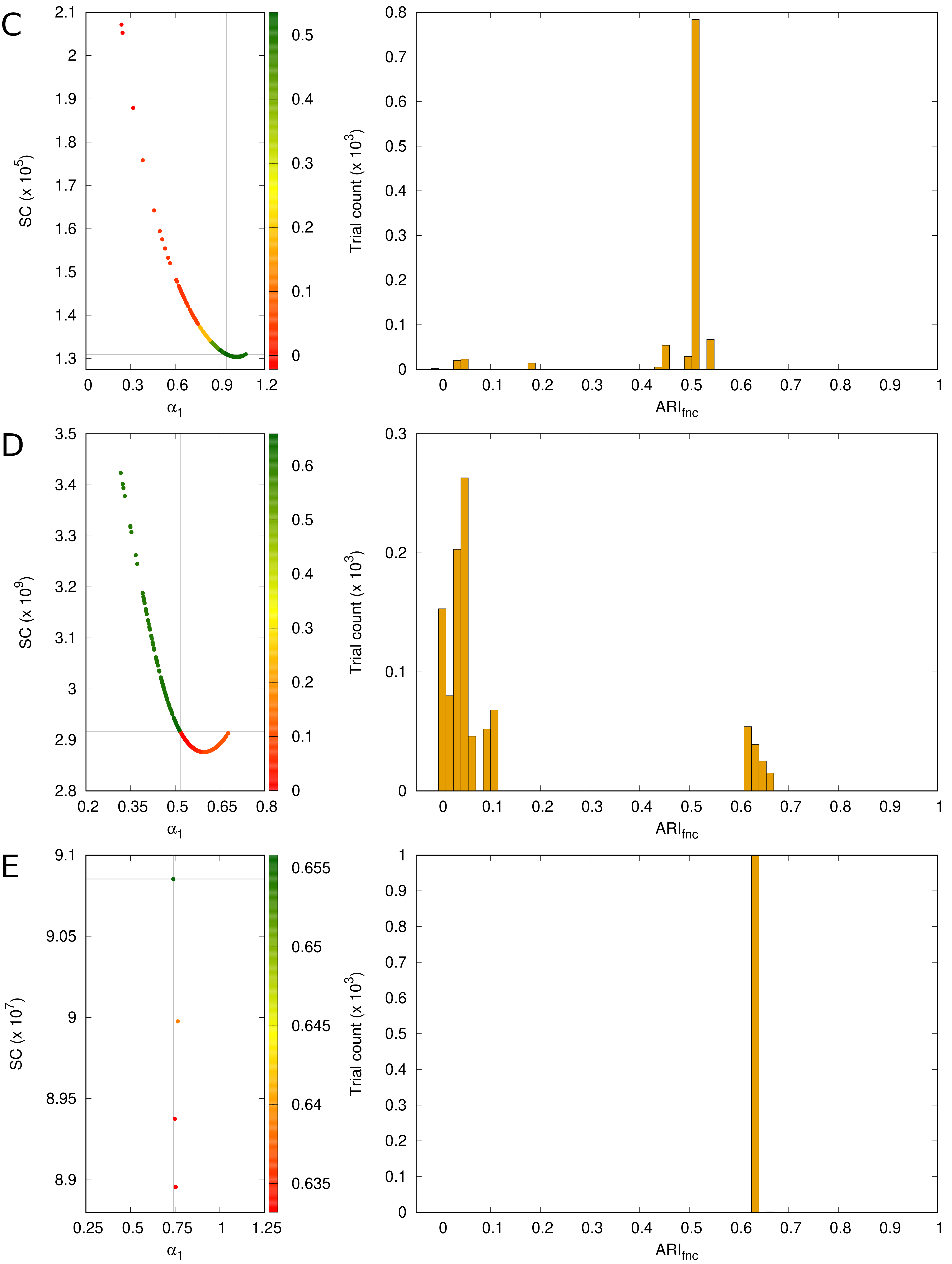}}
\caption{Continued.}
\end{figure}

\section{Discussion}
\label{disc}

Table~\ref{tab2} confirms, for the selection of data sets we are considering,
what by and large has been known for a long while. That is, that scaling by
$1/\sigma_k$ can sometimes be worse than simply attempting to partition the data
in $X$ into clusters without any scaling. In the table, this is the case mainly
of the Iris data set. Table~\ref{tab2} also confirms what has been known since
the recent introduction of scaling by $1/\sigma_k^\mathrm{pool}$, which is that
proceeding in this way, once again in the case of Iris, leads to superior
performance. The table goes farther than this, however, since it also makes
clear that the fallback role of $\sigma_k$ as a surrogate for
$\sigma_k^\mathrm{pool}$ in the approach of \cite{rz20} may be taken more
frequently than initially realized. This is shown in the table for all but the
Iris data set.

But the most relevant contribution of the results in Table~\ref{tab2} is the
realization that in almost all cases the best performing set of $\alpha_k$'s for
each data set performs strictly better than the other three alternatives. The
only exception is the BCW-Diag-10 data set, although in this case every one of
the sets of $\alpha_k$'s can be said to lie, so to speak, in the same ballpark
as $1/\sigma_k$ (or $1/\sigma_k^\mathrm{pool}$). In fact, the plots in
Figure~\ref{cand}(E) strongly suggest that scaling the data for BCW-Diag-10 by
the outcome of virtually any of the random trials with Problem~P would be
equally acceptable. This would be so even if a reference partition (and hence
$\mathrm{ARI_{fnc}}$ values) had not been available, because comparing the
obtained partitions with one another would already suffice.

Of course, the latter is based almost entirely on the highly concentrated
character of the $\mathrm{ARI_{fnc}}$ histogram in Figure~\ref{cand}(E), which
to a degree is also true of Figures~\ref{cand}(B) and~\ref{cand}(C), which
refer to the BCW and BC-DR3 data sets, respectively. For each of these two data
sets, comparing the partitions resulting from the random trials with Problem~P
with one another, and adopting any of those that by the $\mathrm{ARI_{fnc}}$
histogram seem not only to be one and the same but also to recur very frequently
during the trials, would lead to equally acceptable scaling decisions.

This leaves us with the Iris and BNA-DR3 data sets. In these two cases, choosing
the set of $\alpha_k$'s to use out of those produced by the random trials with
Problem~P by simply comparing the obtained partitions and looking for a
consensus with strong support would lead to disastrous results. This is clear
from the $\mathrm{ARI_{fnc}}$ histograms in Figures~\ref{cand}(A)
and~\ref{cand}(D), which peak significantly to the left of the best values
attained in the trials. Beyond comparing obtained partitions, one must therefore
also use one's knowledge of the domain in question and look at what they are
doing to the data. The guiding principle to be used is essentially in the spirit
of our discussion in Section~\ref{scompl}: in the end, the candidate set of
$\alpha_k$'s to be chosen must lead to a partition that makes sense, either
visually or by inspection of the ``midrange'' distances between samples, those
that can be more easily mistaken for intracluster when they are intercluster or
conversely.

We proceed with the aid of Table~\ref{tab3}, which lists each $1/\sigma_k$ and
each $\alpha_k/\sigma_k$ (this one for the highest $\mathrm{ARI_{fnc}}$ listed
in Table~\ref{tab2}) for Iris and BNA-DR3. Note, for the Iris data set, that
$k=1$ and $k=3$ are the dimensions for which switching from scaling by
$1/\sigma_k$ to scaling by $\alpha_k/\sigma_k$ provides the greatest
scaling-factor reduction and amplification (given, in fact, by the value of
$\alpha_k$), respectively. For the BNA-DR3 data set, dimension $k=1$ has its
weight on distances strongly reduced in moving from the former scaling scheme to
the latter, while for both $k=2$ and $k=3$ the scaling factor is amplified by
about the same proportion.

\begin{table}[t]
\centering
\caption{Scaling factors used in Figures~\ref{iris} (Iris) and~\ref{bnote}
(BNA-DR3). The $\alpha_k$'s are the ones leading to the highest values of
$\mathrm{ARI_{fnc}}$ in the intervals on the rightmost column of
Table~\ref{tab2}.}
\label{tab3}
\makebox[\textwidth][c]{\begin{tabular}{lllll}
\hline
$k$ & \multicolumn{2}{c}{Iris} & \multicolumn{2}{c}{BNA-DR3} \\
\cline{2-5}
& $1/\sigma_k$ & $\alpha_k/\sigma_k$ & $1/\sigma_k$ & $\alpha_k/\sigma_k$ \\
\hline
1 & 1.207 & 0.453 & 0.141 & 0.073 \\
2 & 2.294 & 1.291 & 0.327 & 0.373 \\
3 & 0.566 & 0.859 & 0.477 & 0.571 \\
4 & 1.311 & 1.459 & & \\
\hline
\end{tabular}
}
\end{table}

\begin{figure}[p]
\centering
\makebox[\textwidth][c]{\includegraphics[scale=0.35]{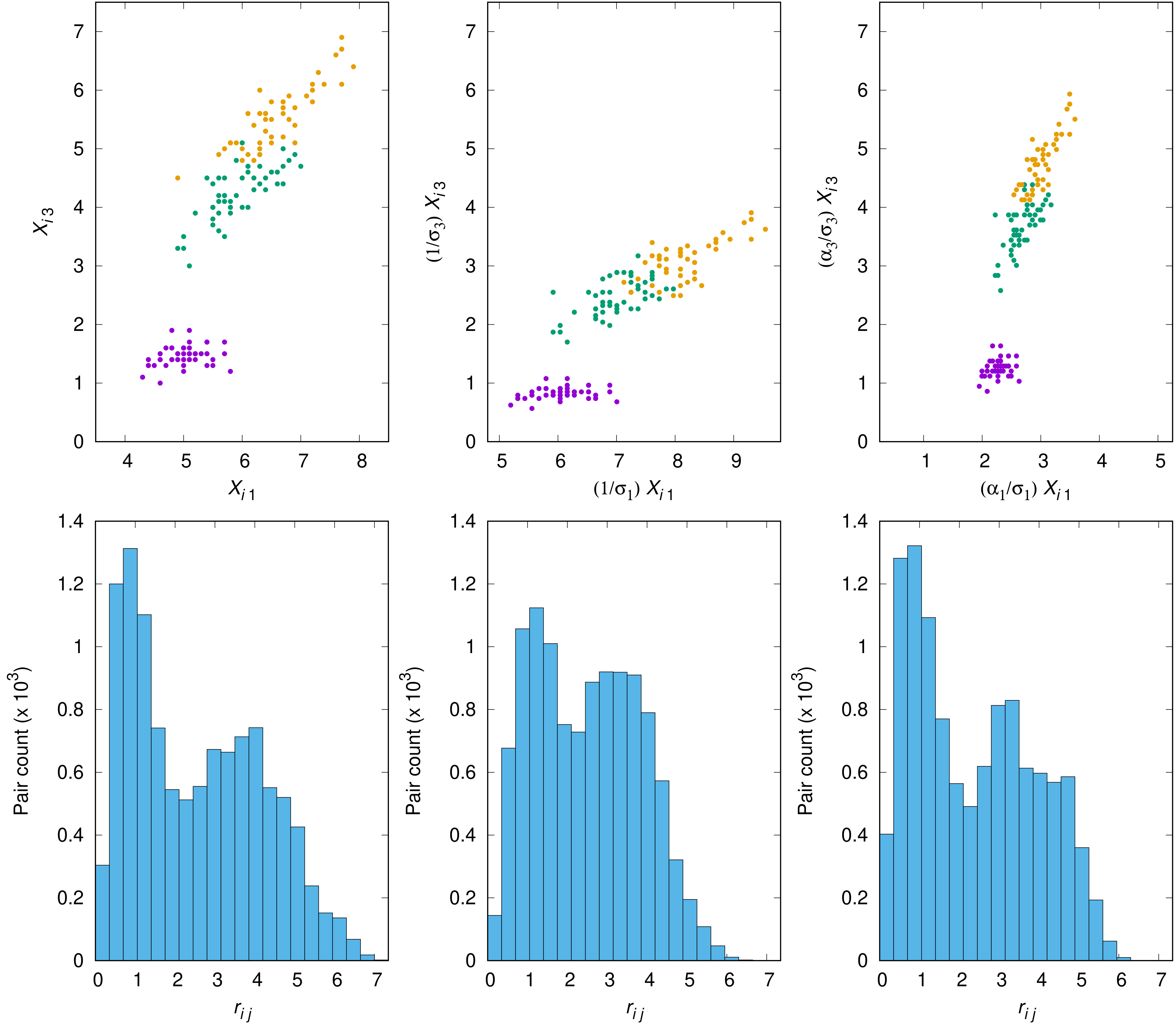}}
\caption{Reference partition for the Iris data set (leftmost column of panels)
and the effects of two scaling schemes: scaling by $1/\sigma_k$ (middle column)
and scaling by $\alpha_k/\sigma_k$ (rightmost column), with factors as in
Table~\ref{tab3}. Effects can be seen both with respect to the shape of the
data set (top row of panels, all plots drawn to the same scale) and to the
distribution of distances between samples (the $r_{ij}$'s; bottom row, all plots
drawn to the same scale).}
\label{iris}
\end{figure}

For the Iris data set, in Figure~\ref{iris} we give six panels. These are
arranged in three columns, the leftmost one dedicated to the data set's
reference partition, each of the other two to a different scaling scheme
(scaling by $1/\sigma_k$ and scaling by $\alpha_k/\sigma_k$, with factors as in
Table~\ref{tab3}). The top panel in each column contains a scatterplot of the
samples, each color-coded for the data set's three classes, as represented by
dimensions $k=1$ and $k=3$ (cf.\ the discussion above in reference to
Table~\ref{tab3}). The bottom panel is a histogram of the $r_{ij}$'s, the
distances between samples in $d$-dimensional real space. It is important to
emphasize that, in regard to the leftmost column, and unlike what happens with
the other two, the scatterplot in it is color-coded to reflect the reference
partition, not the obtained partition that results from the no-scaling scheme.

As we examine the scatterplots in the figure we see that scaling by $1/\sigma_k$
stretches dimension $k=1$ excessively just as dimension $k=3$ is excessively
shrunk, resulting in more confusion between the two clusters that are not
linearly separable. We also see why scaling by $\alpha_k/\sigma_k$ as in
Table~\ref{tab3} is a better choice: the previous stretching of dimension $k=1$
and the shrinking of dimension $k=3$ are both undone, though to different
degrees, which allows some of the previously added confusion to be reverted.
Examining the distance histograms reveals that scaling by $1/\sigma_k$ causes
distances to become more concentrated, lengthening some of the smallest ones and
shortening some of the largest. So another way to view the further confusion
added by this scaling scheme is to recognize that it affects the already
potentially problematic midrange distances. Scaling by $\alpha_k/\sigma_k$
restores the overall appearance of the no-scaling histogram (the one in the
leftmost column), but seemingly with sharper focus around those distances. This
is important because, as we know from Table~\ref{tab2}, scaling the Iris data
set by the $\alpha_k/\sigma_k$ factors of Table~\ref{tab3} improves not only on
the use of the $1/\sigma_k$ factors but also on the no-scaling scheme, on which
k-means already performs more than reasonably well.

\begin{figure}[p]
\centering
\makebox[\textwidth][c]{\includegraphics[scale=0.35]{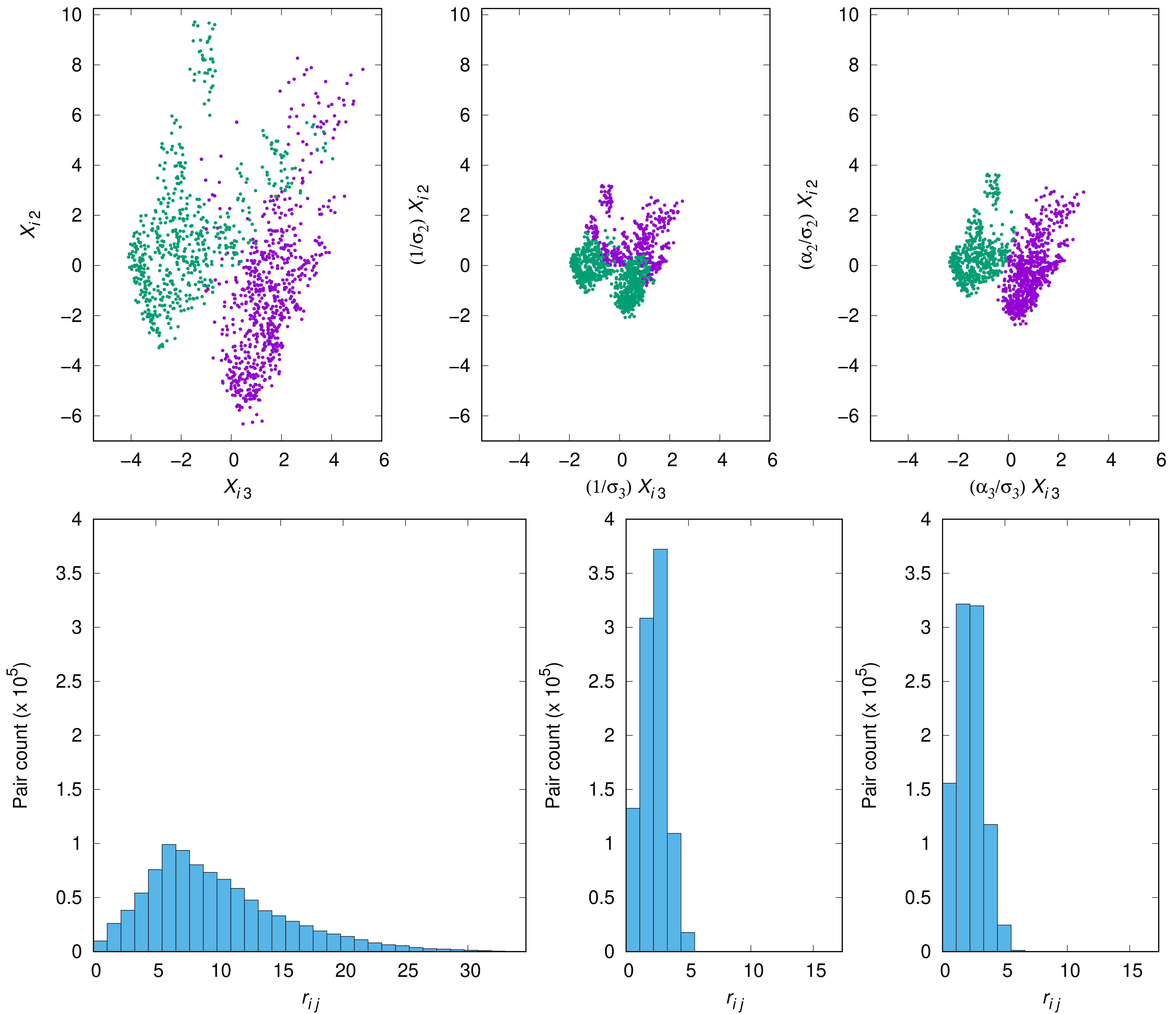}}
\caption{As in Figure~\ref{iris}, now for the BNA-DR3 data set.}
\label{bnote}
\end{figure}

Figure~\ref{bnote} has the same six panels as Figure~\ref{iris}, and also
identically arranged, but now referring to the BNA-DR3 data set. This data set
provides a much more striking contrast between the two scaling schemes given in
Table~\ref{tab3} than Iris, as per Table~\ref{tab2} the ratio of the
$\mathrm{ARI_{fnc}}$ value that scaling by $\alpha_k/\sigma_k$ yields to that
of scaling by $1/\sigma_k$ is about 28.65.

The most direct pictorial evidence we have of this comes from comparing the
middle and rightmost scatterplots of Figure~\ref{bnote} with each other, having
the leftmost one as the reference partition. Because the ratio of
$\alpha_k/\sigma_k$ to $1/\sigma_k$ (i.e., the value of $\alpha_k$) is slightly
higher than only 1.1 for both $k=2$ and $k=3$ (once again, cf.\ our earlier
comment on this), what really accounts for the very significant difference
between the two scaling schemes has to do with dimension $k=1$, for which a
ratio of about 0.517 ensues. Visually inspecting the two obtained partitions
vis-\`a-vis the reference partition provides immediate confirmation of how
crucial this shrinking of dimension $k=1$ is. As with the Iris data set,
inspecting the histograms in the figure provides insight similar to the one we
gleaned in that case. Even though the middle and rightmost histograms,
corresponding respectively to scaling by $1/\sigma_k$ and $\alpha_k/\sigma_k$,
may seem similar to each other particularly when viewed in comparison to the
no-scaling histogram (the leftmost one), closer inspection tells a different
story. That is, moving from scaling by $1/\sigma_k$ to scaling by
$\alpha_k/\sigma_k$ seems to restore some of the no-scaling histogram's slow
descent from its peak through the midrange distances. This comes about by
virtue of both a lower peak and the appearance of some residual pair count
beyond the $r_{ij}=5$ bar when the additional $\alpha_k$ factor is put to use.
This is curious, especially as we note from Table~\ref{tab2} that no scaling
and scaling by $1/\sigma_k$ both lead k-means to essentially the same poor
performance. This seems to be suggesting that, in the no-scaling scheme, such
poor performance is to be attributed essentially to the excessive spread of
distances.

\section{Concluding remarks}
\label{concl}

In this paper we have revisited the problem of scaling a data set's dimensions
to facilitate clustering by those methods that, like k-means, make explicit use
of distances between samples. For each dimension $k$, we have framed our study
as the determination of a scale factor $\alpha_k>0$ to be applied on top of the
customary division by $\sigma_k$, the standard deviation of the data in that
dimension. That is, we have targeted a scaling factor of the form
$\alpha_k/\sigma_k$. Our guiding principle has been to focus on the effects of
scaling the data on the multidimensional shapes that ensue: essentially, we
have equated any facilitation of the clustering task with mistaking
intercluster distances for intracluster distances (or conversely) as seldom
as possible. Because we normally think of the former type of distances as being
large, and the latter as being small, we have aimed our efforts at midrange
distances.

To make such notions precise, we enlisted the shape complexity of the scaled
data, given by $\mathrm{SC}$, which depends heavily on the data matrix $X$ (as a
constant) and on the various $\alpha_k$'s (as variables). The function
$\mathrm{SC}$ embodies a lot of the tension between large and small distances
between samples and, as such, allows midrange distances to be characterized as
being the equilibrium between extremes that occurs at those $\alpha_k$'s for
which the gradient of $\mathrm{SC}$ is zero. We have viewed such scaling-factor
sets as candidates, each one obtained by solving Problem~P, given in
Eqs.~(\ref{obj})--(\ref{c2}), from a randomly chosen initial point.

All our results refer to data sets that are essentially manageable when
considering both their numbers of samples and numbers of dimensions. The
performance of k-means clustering on them, as measured by $\mathrm{ARI_{fnc}}$,
ranges from very poor to well above average, so calling them ``manageable''
refers not at all to how amenable to clustering by k-means they are, but rather
to the possibility of solving Problem~P for them multiple times without too
much computational effort.

Our results can be summarized very simply: for all data sets we tackled,
generating scaling-factor candidate sets via Problem~P has yielded at least one
set for which scaling by $\alpha_k/\sigma_k$ (as opposed to $1/\sigma_k$) leads
to strictly better performance (with one single exception, where ``strictly
better'' becomes ``comparable''). The overall method cannot be used as a blind
procedure, though, since in at least two cases we came across the need for
carefully considered visual inspections of the scaled data, perhaps even of
their distance histograms.

In spite of the radial invariance of $\mathrm{SC}$, which we used to constrain
the $\alpha_k$'s when formulating Problem~P, the number of possibilities outside
the reach of Problem~P is limitless. Suppose, for example, that we use the
following alternative formulation of the nonlinear programming problem.
\begin{align}
\text{maximize }
&\mathrm{SC}\\
\text{subject to }
&\alpha_k\ge 10^{-5}.&\forall k\in\{1,\ldots,d\}
\label{c3}
\end{align}
In significant ways this is still in the spirit of Problem~P, even though it
limits the notion of a gradient-zero point to those that correspond to local
maxima of $\mathrm{SC}$. We mention this particular formulation because solving
it for the Iris data set, as explained in Section~\ref{exp} for Problem~P (now
with \verb|FindMaximum| substituting for \verb|FindMinimum| and selecting the
initial points from $[10^{-5},1]^d$), yielded $\mathrm{ARI_{fnc}}=0.922$ in the
best case, with $\alpha_1=\alpha_2=10^{-5}$, $\alpha_3=5.09372\times 10^{17}$,
and $\alpha_4=2.48504\times 10^{17}$. This is an interesting outcome, and not
only because it surpasses the best result reported in Table~\ref{tab2}
($\mathrm{ARI_{fnc}}=0.904$). What these $\alpha_k$'s are saying is: reduce the
importance of dimensions $k=1$ and $k=2$ as far as allowed by the constraint in
Eq.~(\ref{c3}) while dimensions $k=3$ and $k=4$ are very strongly amplified.
This is to a degree already what Table~\ref{tab3} is suggesting, though only in
relation to scaling by $1/\sigma_k$ and moreover much more timidly.

Alternative formulations like this, and the surprising results they may lead to,
serve to illustrate the rich store of possibilities for shape complexity-based
cluster analysis. Additional investigations to further explore $\mathrm{SC}$ and
its role in helping determine appropriate scaling factors for any given data set
could well be worth the effort. A crucial ingredient will be the use of
techniques to not only reduce the number of dimensions in the data, but also the
number of samples (e.g., as discussed in \cite{ekhg03}), aiming to make possible
the solution of problems like Problem~P. In this regard, we note that, already
for the precursor of the BC-DR3 data set, with $n_\mathrm{orig}=62$ and
$d_\mathrm{orig}=496$, solving one trial with Problem~P is expected to take a
few hundred hours.

\subsection*{Acknowledgments}

The authors acknowledge partial support from Conselho Nacional de
Desenvolvimento Cient\'\i fico e Tecnol\'ogico (CNPq), Coordena\c c\~ao de
Aperfei\c coamento de Pessoal de N\'\i vel Superior (CAPES), and a BBP grant
from Funda\c c\~ao Carlos Chagas Filho de Amparo \`a Pesquisa do Estado do Rio
de Janeiro (FAPERJ).

\bibliography{scompl,scompl-ejaa}
\bibliographystyle{unsrt}

\end{document}